\documentclass[a4paper]{article}

\usepackage{INTERSPEECH2019}
\usepackage{multirow}
\usepackage{amsfonts}

\title{Reverse Transfer Learning: \\
Can Word Embeddings Trained for Different NLP Tasks \\
Improve Neural Language Models?}
\name{Lyan Verwimp$^{1,2}$, Jerome R. Bellegarda$^2$}
\address{
  $^1$ESAT -- PSI, KU Leuven, Belgium \\
  $^2$Apple Inc., USA}
\email{lyan.verwimp@kuleuven.be, jerome@apple.com}

\begin{document}

\maketitle
\begin{abstract}
Natural language processing (NLP) tasks tend to suffer from a paucity of suitably annotated training data, hence the recent success of transfer learning across a wide variety of them. The typical recipe involves: (i) training a deep, possibly bidirectional, neural network with an objective related to language modeling, for which training data is plentiful; and (ii) using the trained network to derive contextual representations that are far richer than standard linear word embeddings such as word2vec, and thus result in important gains. In this work, we wonder whether the opposite perspective is also true: can contextual representations trained for different NLP tasks improve language modeling itself? Since language models (LMs) are predominantly locally optimized, other NLP tasks may help them make better predictions based on the entire semantic fabric of a document. We test the performance of several types of pre-trained embeddings in neural LMs, and we investigate whether it is possible to make the LM more aware of global semantic information through embeddings pre-trained with a domain classification model. Initial experiments suggest that as long as the proper objective criterion is used during training, pre-trained embeddings are likely to be beneficial for neural language modeling. 
\end{abstract}

\section{Introduction}

\label{sec:intro}

Transfer learning involves training a model on one task and re-purposing it for a related task with the aim of improving performance on the second task by exploiting knowledge from the first task. This is expecially true when a smaller amount of data is available for supervised training on the second task. 

In NLP, transfer learning is typically done by using language modeling, inherently a self-annotated task, to improve performance on other NLP tasks, where efficient use of annotated training data is necessary~\cite{elmo,ulmfit,openai-transformer,bert}. 
Usually a deep network is trained with a task related to language modeling 
on large amounts of data,
after which the weights of the network are fine-tuned for a specific NLP task.
Very recently, it has been shown that a large-scale LM can 
even be employed for other NLP tasks in a zero-shot setting~\cite{gpt-2}.
Language modeling being a core task of NLP, contextual representations derived from this core task lead to improvements on related tasks as well.

But could the opposite perspective also be true: can contextual representations trained for different NLP tasks lead to improvements in language modeling itself? Given the prominence of language modeling just exemplified, such perspective can be viewed as ``reverse transfer learning.'' A natural way to proceed is by injecting into the LM pre-trained embeddings optimized for various purposes, instead of training the embeddings jointly with the rest of the network. 
Since LMs are a crucial part in many applications involving the automatic processing of speech and language such as speech recognition, speech synthesis and machine translation, improving the performance of LMs leads to improved performance in these downstream tasks as well.

One can wonder why we need reverse transfer learning: since LMs do not need
labeled data, there is in principle no shortage of training data,
and transfer learning is mainly used in scenarios where there is a limited amount of labeled data.
However, we hypothesize that there might be information that LMs cannot learn themselves,
regardless of the amount of training data, which can be learned through other NLP tasks.
By construction, LMs are optimized to predict the next word. For a large proportion of the training data, the information that is needed to predict the next word can be found in the short-term context, since many dependencies in language are syntactic in nature. Hence, we hypothesize that the LM naturally learns to largely ignore the global semantic context. However, in certain contexts the entire semantic fabric of the text is critical for predicting certain correct content words.
For example, consider the following extract from a Wikipedia article about hurricane Beatriz:\footnote{Extract from WikiText~\cite{wikitext}.}
\begin{quote}
\textit{In Colima and Jalisco, residents were warned of heavy rains that could trigger flooding and mudslides. Schools across both states canceled classes for June 21. In Colima, 236 shelters were opened to the public. Additionally, the Mexican Navy was placed on standby for \textbf{hurricane} relief.}
\end{quote}
In the sentence where the LM needs to predict ~\textit{hurricane}, there is no explicit referral to the fact that this text talks about a natural disaster. Ranking all words according to the probability assigned by a baseline LM, \textit{hurricane} only ends up at position 28,553 out of a vocabulary of 33,279 words. 
It is thus reasonable to hypothesize that making the LM more aware of the global semantic context would increase the probability of \textit{hurricane}. This hypothesis is supported by the fact that explicitly modeling the context improves the LM quality, see for example~\cite{mikolov-context,wang,dieng,jaech}. 

In this paper, we investigate which type of pre-trained embeddings can improve the accuracy of neural LMs. To test the hypothesis formulated earlier, we make a distinction between two classes: (i) embeddings trained with an objective function that is closely related to language modeling, and (ii) embeddings trained with an objective function that promotes global semantic information.
Notice that, in the latter case, we test the possibility to do transfer learning not only from a different NLP task to a LM, but also from a supervised model to an unsupervised model. This might seem counter-intuitive, but as we already mentioned, training in an unsupervised manner can have the drawback that the model learns to neglect less frequent dependencies in the data.

For the first class (i), we take as baseline \textbf{word2vec} embeddings~\cite{word2vec} that are trained based on the local context of the word. Additionally, we look at more sophisticated non-linear embeddings formed from the states of a \textbf{bidirectional LSTM LM}~\cite{lstm,Sun:12}. LSTMs can accommodate longer histories than the usually limited window size of word2vec.
Note that we train the LSTM on sentence level and reset the state at the beginning of each sentence,
because training a bidirectional LSTM across sentence boundaries would require processing the entire text at once.

Global embedding are most commonly generated based on global matrix factorization (see Section~\ref{related} for a short overview), but we choose to keep the underlying architecture in line with the one above and consider a \textbf{bidirectional LSTM trained as domain classifier} for the second class (ii) of pre-trained embeddings.
The LSTM accordingly admits as input a paragraph-size text and as output the semantic domain associated with that text. Using the states of this domain classifier as embeddings in the LM forces the LM to pay attention to the global semantic category/domain of the text. In this case, transfer learning thus operates from a domain classifier to the LM.

In the remainder of the paper, we first describe our approach in section~\ref{approach} and our experimental setup in section~\ref{sec:setup}. Next, we present experimental results in section~\ref{sec:exp} and a conclusion in section~\ref{sec:concl}.

\section{Related work}
\label{related}

\textbf{Pre-trained embeddings} 
Embeddings are commonly used in NLP tasks,
e.g. information retrieval~\cite{Manning}, document classification~\cite{Sebastiani}, question answering~\cite{Tellex}, and neural language modeling~\cite{Bengio}). 
The most elementary embedding is based on 1-of-N or `one-hot' encoding, where every word in an underlying vocabulary of size $N$ is represented by a sparse vector of dimension $N$ (with 1 at the index of the word and 0 elsewhere). One-hot embeddings are usually not used as is but mapped to dense vectors in a lower-dimensional continuous vector space. 
In the course of this dimensionality reduction, the mapping seems to advantageously capture a certain amount of semantic and/or syntactic and/or pragmatic information about the words.

There are two basic classes of continuous-space word embeddings: (i) representations derived from the local context of the word (e.g., the previous $L$ and possibly next $L$ words, where typically $L$ is a small integer), and (ii) representations that exploit the global context surrounding it (e.g., the entire document in which it appears). Methods leveraging global context include global matrix factorization approaches using word-document co-occurrence counts, such as latent semantic analysis (LSA)~\cite{lsa,Bellegarda}, as well as more recent log-linear regression modeling like GloVe which uses word-word co-occurrence counts~\cite{glove}. 
Although the original GloVe paper limits the context for co-occurrences to maximum 10 words, in principle this window size can be increased.
Methods leveraging local context include prediction-based approaches using neural network architectures, such as word2vec continuous bag-of-words and skip-gram models~\cite{word2vec}, the projection layer technique from neural language modeling~\cite{Bengio}, and bottleneck representations using an auto-encoder configuration~\cite{Liou}\footnote{In theory these methods can be implemented using global context, by using a very large window size or feeding the whole corpus to the neural network at once, but in practice this is not feasible.}. 

Global co-occurrence count methods like LSA lead to word representations that can be considered genuine semantic embeddings, because they expose statistical information
which captures semantic concepts conveyed within entire documents. In contrast, typical prediction-based solutions using neural networks only encapsulate semantic relationships to the extent that they manifest themselves within a local window centered around each word (which is all that is used in the prediction). Thus, the resulting embeddings have inherently limited expressive power when it comes to global semantic information. 

\textbf{Transfer learning} In computer vision, transfer learning has been common practice for years. For example, Zeiler and Fergus~\cite{zeiler} show how training a new softmax classifier on an ImageNet~\cite{imagenet} model can improve the results for related datasets, and Donahue et al.~\cite{donahue} use features based on the activation of a deep convolutional network trained for object recognition for other tasks.

In NLP, recently similar results have been obtained. The ELMo embeddings~\cite{elmo} are functions of the input embeddings and hidden states of a deep bidirectional LM
and achieved state-of-the-art accuracies on question answering,
textual entailment, sentiment analysis, semantic role labeling, coreference resolution 
and named entity extraction.
Other work directly fine-tunes the weights of a trained network, instead of learning linear mappings of the weights. 
Howard and Ruder~\cite{ulmfit} train an LSTM LM and propose several optimizations for fine-tuning the network to new tasks, such as discriminative fine-tuning (different learning rates for different layers) and gradual unfreezing (start by fine-tuning the last layer only, then gradually include the other layers). 
Radford et al.~\cite{openai-transformer} train a Transformer~\cite{transformer} LM and fine-tune it with some task-specific input transformations. During fine-tuning they use language modeling as an auxiliary objective in a multi-task learning setup, similar to~\cite{Rei}.
In follow-up work, Radford et al.~\cite{gpt-2} train a much larger Transformer LM on 40GB of webpages and demonstrate that this model achieves state-of-the-art language modeling accuracy on almost all language modeling benchmarks tested, and reasonable accuracy on reading comprehension, summarization and translation in a zero-short setting, so \textit{without} fine-tuning.
The BERT embeddings~\cite{bert} are trained with a bidirectional Transformer LM, which has been trained as a masked LM and a next sentence predictor, and fine-tuned by adding a single layer on top of the Transformer output.

Transfer learning can be interpreted as a context-dependent extension of the concept of word embeddings to all layers of the network, instead of only using the embedding layer. 
Our approach on the other hand, does transfer learning in the opposite direction by trying to improve the LM with an auxiliary task. We do this by pre-training embeddings to incorporate global semantic information, such as the LSA embeddings do, and by comparing those embeddings with local context embeddings.

\section{Pre-trained embeddings in neural language models}
\label{approach}

\subsection{Pre-trained embeddings}
\label{sec:pretr}

In standard LM training, the embedding matrix is initialized randomly and optimized jointly with the rest of the network. The embedding $\mathbf{e}_t$ is the result of multiplying the embedding matrix $\mathbf{E} \in \mathbb{R}^{V \times E}$ ($V$ = vocabulary size and $E$ = embedding size) with the one-hot encoding of the input word at time step $t$, $\mathbf{x}_t \in \mathbb{R}^v$:

\begin{equation}
    \mathbf{e}_t = \mathbf{E}~\mathbf{x}_t
\end{equation}

Using pre-trained embeddings implies replacing the random embedding matrix $\mathbf{E}$ with a pre-trained matrix $\mathbf{E}_{pre}$. The weights in $\mathbf{E}_{pre}$ can remain fixed or can be updated during LM training. 

We experiment with the following pre-training architectures: word2vec and bidirectional LSTMs trained as LM or as domain classifier (see section~\ref{sec:global-sem-emb}).
Pre-training embeddings for an LSTM LM with a bidirectional LSTM LM might seem counter-intuitive, but the bidirectional embeddings might contain additional information that cannot be derived in a unidirectional setting, while in many applications (e.g.\ real-time speech recognition) a unidirectional LM is preferred as final model.

Word2vec generates context-independent word embeddings,
whereas the embeddings extracted from the LM and the domain classifier are context-dependent. 
We adopt the most straightforward approach to generate
context-independent embeddings: we run the LSTM on our dataset, extract the cell state for the current word, and average over all occurrences of the same word in the dataset. Averaging has the advantage that we can pre-compute the embeddings, which saves computations at test time.




\begin{figure}[t]
\centering
\includegraphics[scale=0.4]{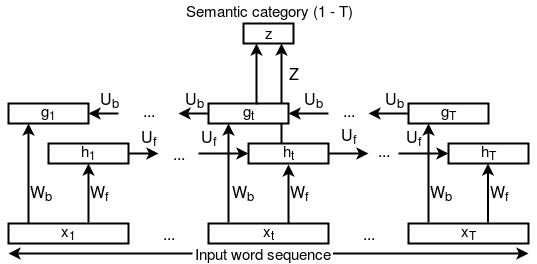}
\caption{Bi-LSTM for training domain embeddings. The rectangles represent LSTM cells.}\label{bi-rnn}
\end{figure}

\subsection{Domain classification embeddings}
\label{sec:global-sem-emb}

Our domain classification model is trained as follows: we assume the current sentence/paragraph/document is composed of $T$ words $x_t$, $1~\leq~t~\leq~T$, and is globally associated with semantic category $z$.  We train a bi-LSTM as depicted in Figure~\ref{bi-rnn} that predicts $z$ for every $x_t$. 
Each word $x$ in the input text is encoded using one-hot encoding, thus $\mathbf{x}_t$ is sparse and its dimension is equal to $V$. The one-hot encoding is mapped to a lower-dimensional continuous embedding $\mathbf{e}_t$ as described in section~\ref{approach}. The output of the domain classifier is calculated as follows:




\begin{equation}
\begin{split}
\mathbf{h}_t &= {\cal T}(\mathbf{W}_f~\mathbf{e}_t + \mathbf{U}_f~\mathbf{h}_{t-1} + \mathbf{b}_f) \\
\mathbf{g}_t &= {\cal T}(\mathbf{W}_b~\mathbf{e}_t + \mathbf{U}_b~\mathbf{g}_{t+1} + \mathbf{b}_b) \\
z &= {\cal S}(\mathbf{Z}~[\mathbf{g}_t \quad \mathbf{h}_t] + \mathbf{b}_o) 
\end{split} \label{eqs}
\end{equation}
where the weight matrices $\mathbf{W}$, $\mathbf{U}$ and $\mathbf{Z}$ and the bias vectors $\mathbf{b}$ are the LSTM parameters, ${\cal T}$ denotes the LSTM equations, and ${\cal S}$ denotes the softmax activation function. We use as word embedding the state of the network $\mathbf{s}_t = [\mathbf{g}_t \quad \mathbf{h}_t]$ (concatenation of forward and backward states). 

This architecture addresses two shortcomings of traditional word2vec-style embeddings.
The first one is the restriction to a local context contained within a finite-sized window centered around each word. By switching to a bi-LSTM, a left and right context of infinite length can in principle be accommodated. This opens up the possibility of handling not just a sentence, but an entire paragraph, or even a full document. The second shortcoming has to do with the prediction target itself. All neural network solutions to date predict either a word in context or the local context itself, which cannot adequately reflect global semantic information. Instead, we predict the semantic domain directly from the input text considered.

\section{Experimental setup}
\label{sec:setup}

We train LSTM LMs consisting of 1 layer of size 128\footnote{All hyperparameters are chosen based on a trade-off between accuracy and available resources.} on an in-house dataset of 10M words and a vocabulary of 25k. The dataset is very diverse in terms of domain and style, containing both formal and informal text. We also train word embeddings on a larger superset of 5B words, containing the same mixture of domains and styles. 
The LSTMs are optimized with Adam~\cite{adam}. 
All experiments in this paper keep the pre-trained weights fixed during training of the LM.
To avoid confusion, we refer to the lower-dimensional mapping of the input before it is given to the LSTM as `compression layer'.
The baseline LM naturally uses a compression layer, otherwise its input would consist of one-hot vectors. The LMs with pre-trained embeddings, for which the input already consists of a continuous vector, can have an additional compression layer. We found that using a compression layer improves results for the LM with domain embeddings, but not for the LM with word2vec embeddings.

The input embeddings are all of size 256 and have been carefully normalized: the best settings were unit length for the word2vec embeddings and mean-variance normalization (we compute the mean and variance for each dimension, and normalize by subtracting the mean and dividing by the variance) for the domain embeddings. Default training parameters are used for the word2vec embeddings. For the bi-LSTMs, concatenating the forward and backward state (each of size 128) yielded better results than averaging them (each of size 256).
As additional baseline, we also considered freely available CBOW embeddings trained on a 100B word dataset from Google News, that are of size 300~\cite{google-word2vec}.

\begin{table*}[h]
\caption{Results for using several types of pre-trained embeddings in neural LMs.}\label{tab:overview}
\centering
\begin{tabular}{ c c c c c }
\toprule
\multirow{2}{*}{\textbf{Pretrained embeddings?}}&\multicolumn{3}{c}{\textbf{Pre-training settings}}&\textbf{PPL} \\
&\textbf{Dataset}&\textbf{Type of data}&\textbf{Training objective}& \\
\midrule
no&/&/&/&189 \\
\midrule
yes&LM superset (5B)&sentences&word2vec&\textbf{162} \\
yes&Google News (100B)&sentences&word2vec&195 \\
\midrule
yes&domain small (7M)&paragraphs&domain classifier&255 \\
yes&domain small (7M)&sentences&domain classifier&239 \\
\midrule
yes&domain large (70M)&paragraphs&word2vec&179 \\
yes&domain large (70M)&paragraphs&domain classifier&245 \\
\midrule
yes&LM data (10M)&sentences&LM&185 \\
\bottomrule
\end{tabular}
\end{table*}

The domain embeddings are trained on a dataset containing paragraphs with associated domain labels (e.g.~sports, technology\ldots). To generate these labels, we performed $k$-means clustering in an initial LSA space and picked the $k$ most dissimilar clusters. We then reclassified the corpus against these $k$ clusters, and adjusted the clusters to include only those documents classified with high confidence. Next, we recomputed the LSA space based only on those documents, and reclassified the entire corpus against the adjusted clusters. After several iterations of this procedure, the final clusters were semantically well separated.
The full dataset contains 70M words, but we also use a subset of 7M to speed up optimization.

\section{Experimental results and discussion}
\label{sec:exp}

In Table~\ref{tab:overview}, we present perplexity (PPL) results for several types of pre-trained embeddings.

Firstly, we observe a 14\% relative perplexity reduction with respect to the baseline by using word2vec embeddings trained on the 5B superset. In contrast, the Google embeddings, trained on a dataset which is 20 times larger than ours, did not improve the quality of the LM, probably because the dataset is quite different in nature and/or is less well normalized. From these results we can conclude that pre-trained embeddings can improve the language modeling quality, but only if training content is aligned with the task at hand. Similar results are reported by Kim~\cite{Kim} and Patel et al.~\cite{Patel}: refining the Google embeddings for the described task and data improves results.


Next, we report the best results for domain classification embeddings pre-trained on the 7M words dataset, and we observe that these embeddings do not improve the PPL. Even though paragraphs inherently contain richer semantic information than sentences, it seems that a match in the format of the training data is some important for the LM, since training the domain classification model on sentences gives a small improvement. 


Upon closer inspection of the domain embeddings, we observed many saturated values, as a result of training on whole paragraphs. Since the state is only reset at the beginning of a new paragraph, state values can easily grow very large or very small. However, we experimented with several methods to deal with this problem, such as different clipping settings and normalizations, and found that results did not materially improve. Thus, we believe that data mismatch, not value saturation, is the main problem preventing a more beneficial contribution from domain embeddings.

After careful optimization on the small domain dataset, we check whether our results extrapolate to the larger domain dataset.
First, we train word2vec embeddings to compare the impact of the training data itself: we observe that the result is better than the baseline LM but worse than for the LM superset because the training data is smaller and matches less well with the LM training data. Note that even though the word2vec embeddings are trained on paragraph data, they can only profit from cross-sentence dependencies as far as those fall within the limited window size used for training. 
Second, we see that there is a large gap between the domain embeddings and the word2vec embeddings. Thus, we hypothesize that congruence with the objective function of the LM---predicting the next word---is particularly critical, given that word2vec embeddings are similarly trained for local prediction, while the semantic embeddings are, in contrast, trained for global prediction.



This hypothesis is further supported by the result in the last row of Table~\ref{tab:overview}, where we show that the embeddings of a bidirectional LM trained on the LM dataset itself give a small improvement with respect to the baseline LM. This is especially true given that this improvement is likely to increase as the embeddings are trained on more data.

\section{Conclusion and future work}
\label{sec:concl}

The purpose of this work was to assess the potential benefits of leveraging word embeddings pre-trained for different NLP tasks in neural LMs. In other words, can we do ``reverse transfer learning'' in favor of the LM itself?
As an initial foray into this line of investigation, we compared embeddings trained with models that are closely related to language modeling, and with a model that is trained to encode global semantic information. Our motivation was that the latter embeddings might counteract the effects of optimizing LMs primarily for local prediction. 
We found, however, that the best results are obtained when training the word embeddings on the same type of data (sentences) with a closely related objective function (word2vec or bi-LM).  In other words, pre-trained embeddings work best when the pre-training task closely resembles the target task, which is in line with previous findings, e.g.~\cite{Zamani,Liu}.



However, we showed that there is promise in using pre-trained embeddings in LMs as long as the objective functions of the embedding model and the LM are congruent: despite having been trained on a much smaller dataset, bidirectional LM embeddings were observed to lead to a meaningful reduction in perplexity. Continuing in that direction, we believe that the natural next step would be to consider a multi-task learning objective. It has already been shown that using language modeling as an auxiliary task can improve the performance of other NLP tasks in a multi-task model~\cite{Rei}. In a similar vein, 
training a LM to simultaneously predict the next word and a semantic label could help the LM to focus more on global semantic information that might be useful for prediction.

\bibliographystyle{IEEEtran}

\bibliography{mybib}

\begin{thebibliography}{10}
\providecommand{\url}[1]{#1}
\csname url@samestyle\endcsname
\providecommand{\newblock}{\relax}
\providecommand{\bibinfo}[2]{#2}
\providecommand{\BIBentrySTDinterwordspacing}{\spaceskip=0pt\relax}
\providecommand{\BIBentryALTinterwordstretchfactor}{4}
\providecommand{\BIBentryALTinterwordspacing}{\spaceskip=\fontdimen2\font plus
\BIBentryALTinterwordstretchfactor\fontdimen3\font minus
  \fontdimen4\font\relax}
\providecommand{\BIBforeignlanguage}[2]{{%
\expandafter\ifx\csname l@#1\endcsname\relax
\typeout{** WARNING: IEEEtran.bst: No hyphenation pattern has been}%
\typeout{** loaded for the language `#1'. Using the pattern for}%
\typeout{** the default language instead.}%
\else
\language=\csname l@#1\endcsname
\fi
#2}}
\providecommand{\BIBdecl}{\relax}
\BIBdecl

\bibitem{elmo}
M.~E. Peters, M.~Neumann, M.~Iyyer, M.~Gardner, C.~Clark, K.~Lee, and
  L.~Zettlemoyer, ``{Deep contextualized word representations},'' in
  \emph{Proceedings of NAACL-HLT}, 2018, pp. 2227--2237.

\bibitem{ulmfit}
J.~Howard and S.~Ruder, ``{Universal Language Model Fine-tuning for Text
  Classification},'' in \emph{Proceedings of the Annual Meeting of the
  Association for Computational Linguistics (ACL)}, 2018, pp. 328--339.

\bibitem{openai-transformer}
A.~Radford, K.~Narasimhan, T.~Salimans, and I.~Sutskever, ``{Improving language
  understanding with unsupervised learning},'' {OpenAI}, Tech. Rep., 2018.

\bibitem{bert}
J.~Devlin, M.-W. Chang, K.~Lee, and K.~Toutanova, ``{BERT: Pre-training of Deep
  Bidirectional Transformers for Language Understanding},''
  \emph{arXiv:1810.04805}, 2018.

\bibitem{gpt-2}
A.~Radford, J.~Wu, R.~Child, D.~Luan, D.~Amodei, and I.~Sutskever, ``{Language
  Models are Unsupervised Multitask Learners},'' 2019, preprint on webpage at
  \url{blog.openai.com/better-language-models} (OpenAI).

\bibitem{wikitext}
S.~Merity, C.~Xiong, J.~Bradbury, and R.~Socher, ``{Pointer Sentinel Mixture
  Models},'' in \emph{Proceedings of the International Conference for Learning
  Representations (ICLR)}, 2017.

\bibitem{mikolov-context}
T.~Mikolov and G.~Zweig, ``{Context dependent recurrent neural network language
  model},'' in \emph{Proceedings of IEEE Worskhop on Spoken Language Technology
  (SLT)}, 2012, pp. 234--239.

\bibitem{wang}
T.~Wang and K.~Cho, ``{Larger-Context Language Modelling with Recurrent Neural
  Network},'' in \emph{Proceedings of the Annual Meeting of the Association for
  Computational Linguistics (ACL)}, 2016, pp. 1319--1329.

\bibitem{dieng}
A.~B. Dieng, C.~Wang, J.~Gao, and J.~Paisley, ``{TopicRNN: A recurrent neural
  network with long-range semantic dependency},'' in \emph{Proceedings of the
  International Conference on Learning Representations (ICLR)}, 2017.

\bibitem{jaech}
A.~Jaech and M.~Ostendorf, ``{Low-Rank RNN Adaptation for Context-Aware
  Language Modeling},'' \emph{Transactions of the Association for Computational
  Linguistics (TACL)}, vol.~6, pp. 497--510, 2018.

\bibitem{word2vec}
T.~Mikolov, I.~Sutskever, K.~Chen, G.~S. Corrado, and J.~Dean, ``{Distributed
  Representations of Words and Phrases and their Compositionality},'' in
  \emph{Advances in Neural Information Processing Systems (NIPS)}, 2013, pp.
  3111--3119.

\bibitem{lstm}
S.~Hochreiter and J.~Schmidhuber, ``Long short-term memory,'' \emph{Neural
  Computation}, vol.~9, no.~8, pp. 1735--1780, 1997.

\bibitem{Sun:12}
M.~Sundermeyer, R.~Schl\"{u}ter, and H.~Ney, ``{LSTM} {N}eural {N}etworks for
  {L}anguage {M}odeling,'' \emph{Proceedings Interspeech}, pp. 1724--1734,
  2012.

\bibitem{Manning}
C.~D. Manning, P.~Raghavan, and H.~Sch\"{u}tze, \emph{{Introduction to
  Information Retrieval}}.\hskip 1em plus 0.5em minus 0.4em\relax New York,
  USA: Cambridge University Press, 2008.

\bibitem{Sebastiani}
F.~Sebastiani, ``{Machine Learning in Automated Text Categorization},''
  \emph{ACM Computing Surveys (CSUR)}, vol.~34, no.~1, pp. 1--47, 2002.

\bibitem{Tellex}
S.~Tellex, B.~Katz, J.~Lin, A.~Fernandes, and G.~Marton, ``{Quantitative
  Evaluation of Passage Retrieval Algorithms for Question Answering},'' in
  \emph{Proceedings of the 26th annual international ACM SIGIR conference on
  Research and development in information retrieval}, 2003, pp. 41--47.

\bibitem{Bengio}
Y.~Bengio, R.~Ducharme, P.~Vincent, and C.~Jauvin, ``{A Neural Probabilistic
  Language Model},'' \emph{Journal of Machine Learning Research}, vol.~3, pp.
  1137--1155, 2003.

\bibitem{lsa}
S.~Deerwester, S.~T. Dumais, G.~W. Furnas, T.~K. Landauer, and R.~Harshman,
  ``Indexing by latent semantic analysis,'' \emph{Journal of the American
  Society for Information Science}, vol.~41, no.~6, pp. 391--407, 1990.

\bibitem{Bellegarda}
J.~R. Bellegarda, ``{Exploiting Latent Semantic Information in Statistical
  Language Modeling},'' \emph{Proceedings of the IEEE}, vol.~88, no.~8, pp.
  1279--1296, 2000.

\bibitem{glove}
J.~Pennington, R.~Socher, and C.~D. Manning, ``{GloVe: Global Vectors for Word
  Representation},'' in \emph{Proceedings of the Conference on Empirical
  Methods in Natural Language Processing (EMNLP)}, 2014, pp. 1532–--1543.

\bibitem{Liou}
C.-Y. Liou, W.-C. Cheng, J.-W. Liou, and D.-R. Liou, ``{Autoencoder for
  Words},'' \emph{Neurocomputing}, vol. 139, pp. 86--96, 2014.

\bibitem{zeiler}
M.~D. Zeiler and R.~Fergus, ``{Visualizing and Understanding Convolutional
  Networks},'' \emph{Computer Vision -- ECCV 2014}, pp. 813--833, 2014.

\bibitem{imagenet}
A.~Krizhevsky, I.~Sutskever, and G.~E. Hinton, ``{Imagenet classification with
  deep convolutional neural networks},'' in \emph{Advances in Neural
  Information Processing Systems (NIPS)}, 2012.

\bibitem{donahue}
J.~Donahue, Y.~Jia, O.~Vinyals, J.~Hoffman, N.~Zhang, E.~Tzeng, and T.~Darrell,
  ``{DeCAF: A Deep Convolutional Activation Feature for Generic Visual
  Recognition},'' in \emph{Proceedings of the IEEE Conference on Computer
  Vision and Pattern Recognition}, 2014, pp. 647--655.

\bibitem{transformer}
A.~Vaswani, N.~Shazeer, N.~Parmar, J.~Uszkoreit, L.~Jones, A.~N. Gomez,
  L.~Kaiser, and I.~Polosukhin, ``{Attention is all you need},'' \emph{Advances
  in Neural Information Processing Systems (NIPS)}, pp. 6000--6010, 2017.

\bibitem{Rei}
M.~Rei, ``{Semi-supervised Multitask Learning for Sequence Labeling},'' in
  \emph{Proceedings of the Annual Meeting of the Association for Computational
  Linguistics (ACL)}, 2017, pp. 2121--2130.

\bibitem{adam}
D.~P. Kingma and J.~Ba, ``{Adam: A Method for Stochastic Optimization},'' in
  \emph{Proceedings of the International Conference for Learning
  Representations (ICLR)}, 2015.

\bibitem{google-word2vec}
\BIBentryALTinterwordspacing
Google, ``{Google News pre-trained vectors},'' 2013. [Online]. Available:
  \url{https://code.google.com/archive/p/word2vec/}
\BIBentrySTDinterwordspacing

\bibitem{Kim}
Y.~Kim, ``{Convolutional Neural Networks for Sentence Classification},'' in
  \emph{Proceedings of the Conference on Empirical Methods in Natural Language
  Processing (EMNLP)}, 2014, pp. 1746--1751.

\bibitem{Patel}
K.~Patel, D.~Patel, M.~Golakiya, P.~Bhattacharyya, and N.~Birari, ``{Adapting
  Pre-trained Word Embeddings For Use In Medical Coding},'' in
  \emph{Proceedings of the BioNLP 2017 workshop}, 2017, pp. 302--306.

\bibitem{Zamani}
H.~Zamani and W.~B. Croft, ``{Relevance-based Word Embedding},'' in
  \emph{{Proceedings of Special Interest Group on Information Retrieval
  (SIGIR)}}, 2017, pp. 505--514.

\bibitem{Liu}
Q.~Liu, H.~Huang, Y.~Gao, X.~Wei, Y.~Tian, and L.~Liu, ``{Task-oriented Word
  Embedding for Text Classification},'' in \emph{{Proceedings of the 27th
  International Conference on Computational Linguistics (COLING)}}, 2018, pp.
  2023--2032.

\end{thebibliography}

\end{document}